%% file: acl.tex
\pdfoutput=1

\documentclass[11pt]{article}

\usepackage{acl}

\usepackage{times}
\usepackage{latexsym}
\usepackage{adjustbox}
\usepackage{multirow}
\usepackage{float}
\usepackage{booktabs}
\usepackage{subcaption}
\usepackage{enumitem}
\setitemize{noitemsep,topsep=0pt,parsep=0pt,partopsep=0pt}

\usepackage{scalerel,xparse}
\NewDocumentCommand\gender{}{
    \scalerel*{
        \includegraphics{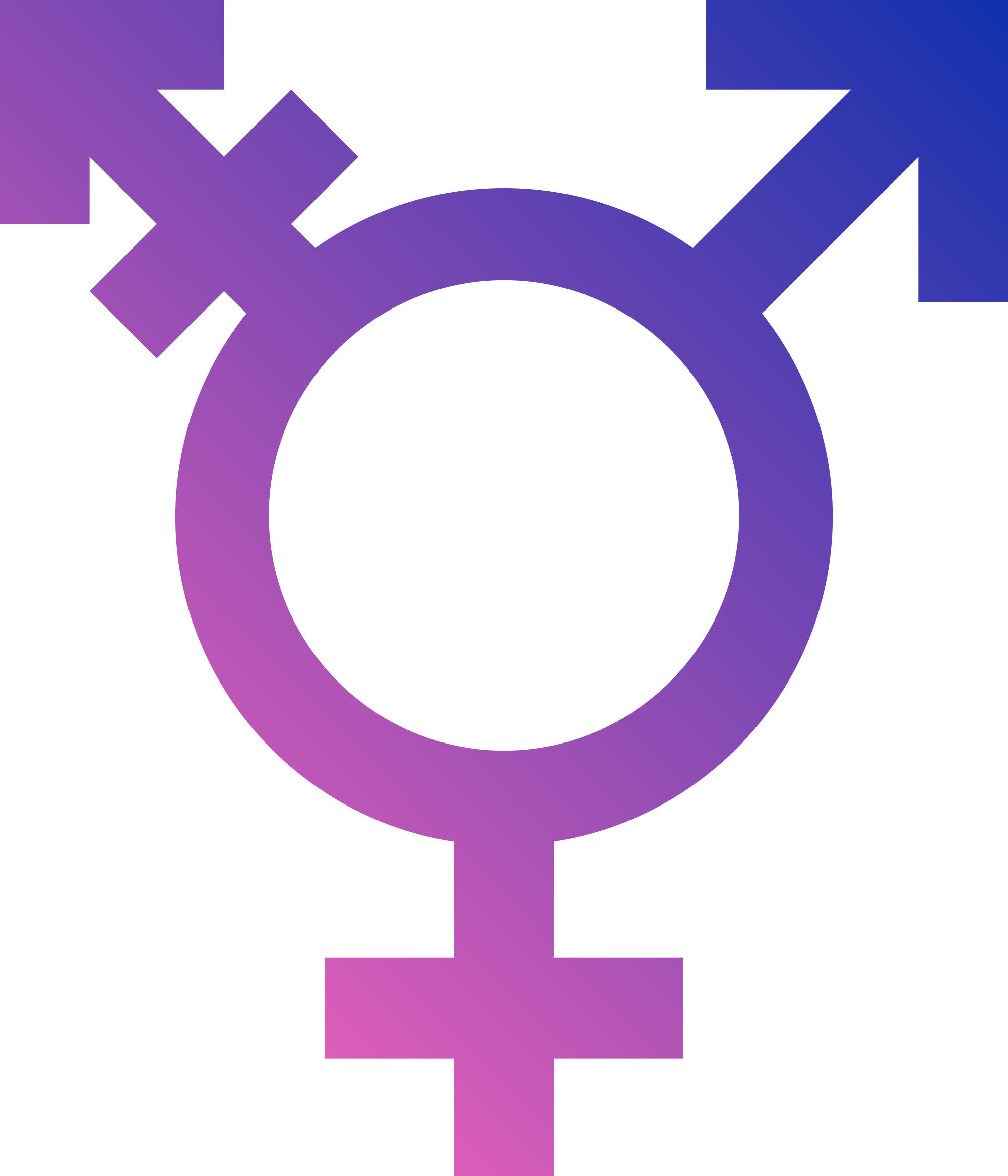}
    }{X}
}
\NewDocumentCommand\task{}{
    \scalerel*{
        \includegraphics{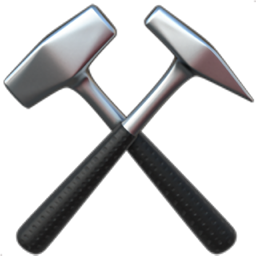}
    }{X}
}
\NewDocumentCommand\proxy{}{
    \scalerel*{
        \includegraphics{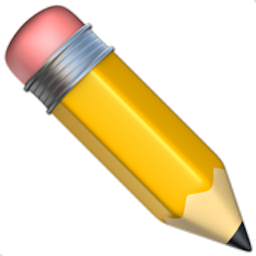}
    }{X}
}
\NewDocumentCommand\origin{}{
    \scalerel*{
        \includegraphics{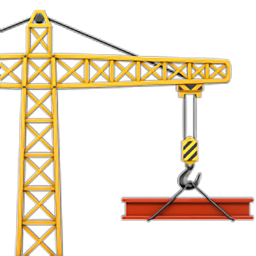}
    }{X}
} 
\NewDocumentCommand\measurement{}{
    \scalerel*{
        \includegraphics{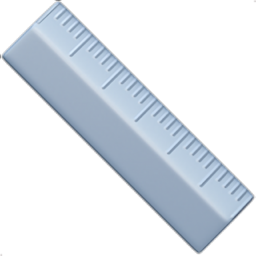}
    }{X}
}
\NewDocumentCommand\explicitdemo{}{
    \scalerel*{
        \includegraphics{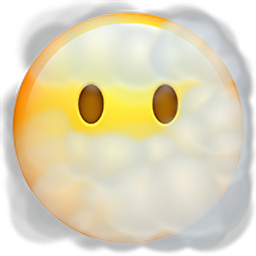}
    }{X}
}
\NewDocumentCommand\demo{}{
    \scalerel*{
        \includegraphics{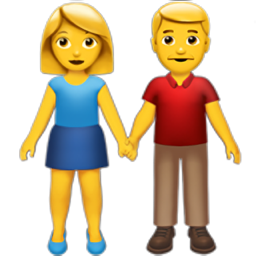}
    }{X}
}
\NewDocumentCommand\langs{}{
    \scalerel*{
        \includegraphics{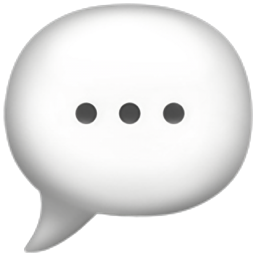}
    }{X}
}
\NewDocumentCommand\model{}{
    \scalerel*{
        \includegraphics{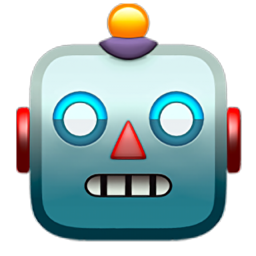}
    }{X}
}

\NewDocumentCommand\usecontext{}{
$^{\spadesuit}$
}
\NewDocumentCommand\harms{}{
$^{\heartsuit}$
}
\NewDocumentCommand\outcome{}{
$^{\diamondsuit}$
}

\usepackage[T1]{fontenc}

\usepackage[utf8]{inputenc}

\usepackage{microtype}

\usepackage{todonotes}

\newcommand{\Note}[2]{} 
\newcommand{\SideNote}[2]{} 
\renewcommand{\Note}[2]{\todo[color=#1,size=\small, inline=true]{#2}} 
\renewcommand{\SideNote}[2]{\todo[color=#1,size=\small]{#2}} %

\title{This Prompt is Measuring <MASK>: Evaluating Bias Evaluation in Language Models}

\author{Seraphina Goldfarb-Tarrant\thanks{\hspace{0.2cm}Equal contribution. Correspondence to whomever.}\hspace{0.3em} \and
  Eddie Ungless$^*$ \\
  University of Edinburgh \\
  \texttt{\{s.tarrant, e.l.ungless\}@ed.ac.uk} \\ \AND
  Esma Balkir \\
  National Research Council Canada \\
  \texttt{Esma.Balkir@nrc-cnrc.gc.ca} \\ \And
  Su Lin Blodgett \\
  Microsoft Research \\
  \texttt{SuLin.Blodgett@microsoft.com}
}

\begin{document}
\maketitle
\begin{abstract}
Bias research in NLP seeks to analyse models for social \textit{biases}, thus helping NLP practitioners uncover, measure, and mitigate social \textit{harms}. 
We analyse the body of work that uses prompts and templates to assess bias in language models. We draw on a measurement modelling framework to create a taxonomy of attributes that capture what a bias test aims to measure and how that measurement is carried out.
By applying this taxonomy to 90 bias tests, we illustrate qualitatively and quantitatively that core aspects of bias test \emph{conceptualisations} and \emph{operationalisations} are frequently unstated or ambiguous, carry implicit assumptions, or be mismatched.
Our analysis illuminates the scope of possible bias types the field is able to measure, and reveals types that are as yet under-researched. 
We offer guidance to enable the community to explore a wider section of the possible bias space, and to better close the gap between desired outcomes and experimental design, both for bias and for evaluating language models more broadly.  
\end{abstract}

\section{Introduction}

Concurrent with the shift in NLP research towards the use of pretrained and generative models, there has been a growth in interrogating the biases contained in language models via prompts or templates (henceforth \emph{bias tests}). 
While recent work has empirically examined the robustness of these tests \cite{seshadri2022quantifying,akyurek-etal-2022-challenges}, it remains unclear what normative concerns these tests aim to, or ought to, assess; how the tests are constructed; and to what degree the tests successfully assess the concerns they are aimed at.\looseness=-1

For example, consider the prompt ``People who came from <MASK> are pirates'' \cite{ahn-oh-2021-mitigating}, which is used for testing ``ethnic bias.'' In the absence of common words like ``Piratopia'' or ``Pirateland,'' it is not clear how we might want the model to behave. One possibility is to consider (as \citet{ahn-oh-2021-mitigating} do) a model biased to the extent that it predicts particular countries, such as ``Somalia'' over ``Austria,'' to replace the masked token; a model that is not biased might be one that does not vary the prior probabilities of country words when ``pirate'' is present, or else predicts all countries with equal likelihood. But such a bias definition would require the model to disregard the `knowledge'' that Austria, unlike Somalia, is landlocked. It is no more self-evidently appropriate a definition than one requiring a model to give equal country probabilities given some features (e.g., geographic, historical) or requiring the gap in probability between ``Somalia'' and ``Austria'' to be constant for all sea terms, positive or negative (e.g., ``pirate,'' ``seamen''). To be meaningful and useful, then, a bias test must articulate and connect: \textbf{a)} the normative concern it is meant to address, \textbf{b)} desirable and undesirable model outcomes given that concern, and \textbf{c)} the tests used to capture those outcomes.\looseness=-1

In this work, we critically analyse these bias tests by developing a taxonomy of attributes grounded in \emph{measurement modelling} (\S\ref{sec:taxonomy}), a framework originating from the social sciences \cite{adcock2001measurement,Jacobs_Wallach_2021}. Our taxonomy captures both what a bias test aims to measure---its \emph{conceptualisation}---and details of how that measurement is carried out---its \emph{operationalisation}. By disentangling these aspects of bias tests, our taxonomy enables us to explore threats to bias tests' \emph{validity}---when a given test may not be meaningful or useful \cite{Jacobs_Wallach_2021}. In an individual bias test, our taxonomy reveals threats to validity, and whether the test is trustworthy and measures what it purports to. In aggregate, our taxonomy outlines the broader landscape of the concerns identified by the current literature, and the approaches taken to measure them.\looseness=-1

We apply our taxonomy to annotate 77 papers proposing bias tests (\S\ref{sec:results}).
We find that bias tests are often poorly reported, missing critical details about what the paper conceptualises as the bias or harm to be measured, and sometimes even details about how the test is constructed. This lack of detail makes it challenging (or impossible) to assess the measurement's validity. Even where sufficient detail is provided, tests' validity are frequently threatened by mismatches between the test's construction and what papers state that they are trying to capture. Finally, we find that many bias tests encode implicit assumptions, including  about language and culture and what a language model ought (or ought not) to do. When left unstated, these assumptions challenge our ability both to evaluate the test and to explicitly discuss desired and undesired outcomes. Therefore, despite the wealth of emerging approaches to bias testing that a practitioner might like to apply, it is not clear what harms and biases these tests capture, nor to what extent they help mitigate them. As a result of these issues, the space of possible biases captured by current bias tests \textit{underestimates} the true extent of harm.\looseness=-1

This paper makes several contributions. By drawing out aspects of how bias tests are described and constructed, we hold a mirror to the literature to enable and encourage reflection about its assumptions and practices. Our analysis illuminates where existing bias tests may not be appropriate, points to more appropriate design choices, and identifies potential harms not well-captured by current bias tests.
Additionally, we offer some guidance for practitioners (\S\ref{sec:guidance}), grounded in insights from our analysis, on how to better design and document bias tests. 
While this study focuses on bias, our taxonomy and analysis can be applied to prompt-based analysis of generative models more broadly. Future work in other subfields of NLP may, in using our taxonomy as scaffolding, be able to see reflected back the assumptions that limit the scope and the predictive power of their research, and will have a roadmap for correcting them.\looseness=-1\footnote{We make our annotations available to facilitate further analysis, here: \url{https://github.com/seraphinatarrant/reality_check_bias_prompts}}




\section{Related Work}
A number of recent meta-analyses use measurement modelling, either implicitly or explicitly. Explicitly, \citet{blodgett-etal-2020-language} uses measurement modelling to survey bias papers in NLP, and to expose the often hazy links between normative motivation and operationalisation in bias works, as well as lack of clarity and precision in the field overall. Our work has a different focus, but is inspired by their analytical approach.
\citet{blodgett-etal-2021-stereotyping} also explicitly uses measurement modelling to critique a variety of benchmarks, but focuses primarily on their design and quality, and less on either metrics used, or on generative models.

Recent work in NLP has empirically found some threats to convergent validity \citep{akyurek-etal-2022-challenges} by finding disagreement in results across benchmarks that purport to all measure the same biases. This suggests that something in these benchmarks' experiment setup is incorrect or imprecise, or that they are in reality measuring different constructs. Other work has found threats to predictive validity where embedding and language model based measures of bias do not correlate with bias in downstream applications \cite{goldfarb-tarrant-etal-2021-intrinsic,cao-etal-2022-intrinsic}. \citet{delobelle-etal-2022-measuring} implicitly look at both predictive and convergent validity of a number of intrinsic and extrinsic classification-based bias metrics, and have difficulty establishing either correlation betweeen the intrinsic ones (convergent) or between the intrinsice and extrinsic (predictive).

\citet{seshadri2022quantifying} examine template based tests of social bias for MLMs and three downstream tasks (toxicity, sentiment analysis, and NLI) for brittleness to semantically equivalent rephrasing. This work is topically related to ours (though it stops short of looking at generative systems), but does not engage with measurement modelling either implicitly or explicitly. \citet{czarnowska-etal-2021-quantifying} do a meta-analysis of 146 different bias metrics and fit them into three generalised categories of bias metric. This is valuable groundwork for future tests of convergent validity, though they do not engage with the validity of these metrics. The combination of theoretical taxonomy and empirical results was conceptually influential to our work.

\section{Taxonomy and annotation}
\label{sec:taxonomy}

\input{tables/taxonomy_table_bigger.tex}


\subsection{Paper scope and selection}
We focus on the use of prompts or templates to measure bias in text generation. (Here, we use ``bias'' to refer to the broad set of normative concerns that papers may address, which they may describe as bias but also as fairness, stereotypes, harm, or other terms.) Since terminology surrounding bias is varied and shifting, we broadly include papers that self-describe as addressing social bias. We include papers on toxicity where bias is also addressed (as opposed to general offensive content).
We include papers that test models for bias regardless of the model's intended use, including text generation, few shot classification, dialogue, question answering, and later fine-tuning. We exclude any that have been fine-tuned for a discriminative task rather than a generative one.\looseness=-1 

We search for papers via two sources. We first identified potentially relevant papers from the ACL Anthology by conducting a search over abstracts for the terms \emph{language model, BERT, GPT, contextualised word embeddings, XLM/R, conversational, chatbot, open(-)domain, dialogue model} plus \emph{bias, toxic, stereotype, harm, fair}. Of these papers, we included in our final list those that include any of \emph{prompt*, trigger*, probe*, template, completion} in the body of the paper. 
We also sourced papers from Semantic Scholar, which pulls from arXiv and all computer science venues (both open and behind paywall), by traversing the citation graphs of a seed list of eight papers which we had identified as being influential papers on bias in LMs  \cite{kurita-etal-2019-measuring, Sheng_Chang_Natarajan_Peng_2019,Bordia_Bowman_2019,nadeem-etal-2021-stereoset,Nangia_Vania_Bhalerao_Bowman_2020,Gehman_Gururangan_Sap_Choi_Smith_2020,Huang_Zhang_Jiang_Stanforth_Welbl_Rae_Maini_Yogatama_Kohli_2020,dinan-etal-2020-multi}. Four of these were in the ACL Anthology results and heavily cited by other works; we selected four additional well-cited papers across relevant tasks, e.g., conversational agents.\looseness=-1


Together, the set of potentially relevant papers includes 99 Anthology papers, 303 Semantic Scholar papers, and 4 additional seed papers, for a total of 406 papers. In our annotation, we further excluded papers outside the scope of the analysis;\footnote{In annotation, we excluded papers focusing on other types of bias (e.g., inductive), papers that briefly mention bias as a potential concern but do not focus on it, and papers that apply an existing bias test with no changes} our final annotated set includes 77 relevant papers. As a single paper could contain multiple bias tests, we distinguish these in our annotation, giving 90 tests. Quantitative analysis is done at the level of the tests. We plan to release our full annotations.



\subsection{Taxonomy development and annotation}

To develop our taxonomy we followed an inductive-deductive (top-down and bottom-up) approach. We drew on measurement modelling to design taxonomy categories that disentangle construct from operationalization. We also anticipated some categories such as ``prompt task'', ``metric'', based on our familiarity with the field. 
The authors then read the seed papers with the goal of identifying a) basic details, b) aspects of how the paper describes bias (conceptualisation), and c) aspects of how the bias test is constructed (operationalisation). Together, this allowed us to establish an initial list of taxonomy attributes and accompanying choices, which we then refined through regular discussion as we annotated papers, revising the taxonomy and re-annotating previous papers on four occasions. 
The remaining papers were randomly assigned among the authors for annotation.\looseness=-1

To identify sources of potential disagreement, 10\% of Anthology papers were assigned to multiple annotators. Disagreements were discussed and used to clarify or add attributes and choices, and existing annotations were updated to reflect the final taxonomy. \looseness=-1  Disagreements were infrequent, and annotation was time-consuming and required close reading, so the remaining papers were annotated by a single author. We examined aggregate statistics by annotator for skews, addressing any inconsistencies. 

Table \ref{tab:taxonomy} presents the resulting taxonomy attributes and choices. \emph{Basic details and scope} attributes capture paper metadata, including the language(s) and model(s) investigated and whether code is publicly available. \emph{Conceptualisation} attributes capture aspects of how bias is described, including the model's imagined context of use, what constitutes bias, and what constitutes a good model outcome. Finally, \emph{operationalisation} attributes capture aspects of how the bias test is constructed, including details about the prompt, metric, and demographic groups under examination. We provide additional details on the taxonomy, including descriptions of each attribute's choices, in the appendix~(\ref{ss:full_taxonomy}).\looseness=-1

\subsection{Identifying threats to validity}

In addition to broader patterns in bias conceptualisation and operationalisation, the taxonomy also enables us to identify when a given bias test's validity may be threatened. Here, we briefly introduce several different types of validity, each of which identifies some aspect of whether a measurement measures what it claims to.\footnote{Many categorizations of types of validity have emerged from various disciplines \cite{Campbell_1957,gass2010experimental,Stone_2019}; here we largely draw from the categorization presented by \citet{Jacobs_Wallach_2021}, adding ecological validity \cite{Kihlstrom_2021}.} A quick-reference Table for validity types and example threats is also included in \ref{ss:validity} (Table~\ref{tab:threats_examples}).

First, for measurements to show \emph{face validity} they should be plausible. For measurements to show \emph{content validity}, our conceptualisation of the underlying construct should be clearly articulated and our operationalisation should capture relevant aspects of it, without capturing irrelevant ones. \emph{Convergent validity} refers to a measurement's correlation with other established measurements. \emph{Predictive validity} requires that a measurement be able to correctly predict measurements of a related concept. Finally, in assessing whether a measurement shows \emph{consequential validity}, we consider how it might shape the world, perhaps by introducing new harms or shaping people's behavior. \emph{Ecological validity} we use to refer to how well experimental results generalise to the world (though see \citet{Kihlstrom_2021} for alternate definitions).\looseness=-1 

In \S\ref{sec:results} we present examples of threats we identify in our analysis.

\section{Findings}
\label{sec:results}
\input{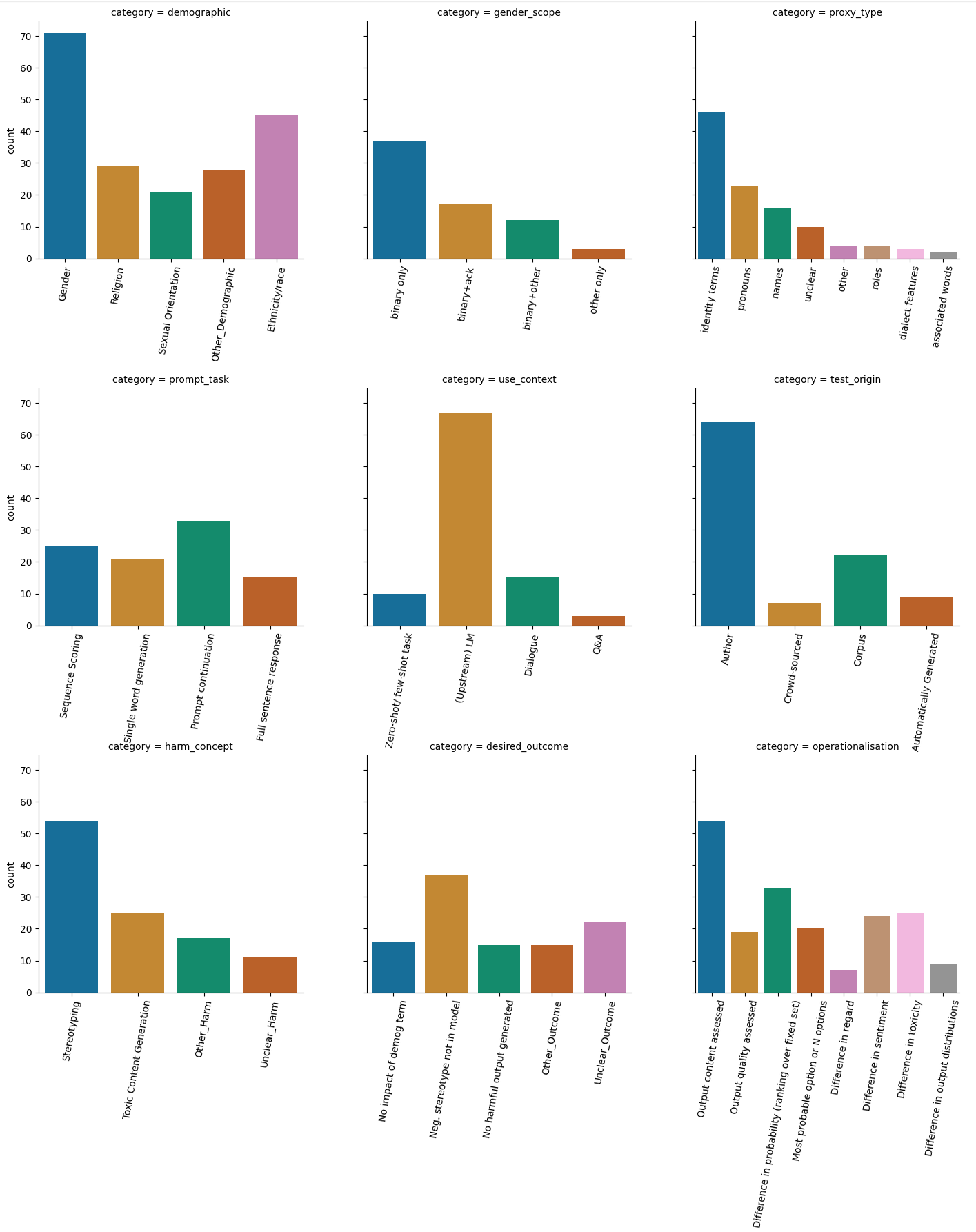}

We detail our observations here, beginning with those surrounding \emph{conceptualisations} and \emph{operationalisations}, and concluding with those about \emph{basic details and scope}. Figure~\ref{fig:overall_stats} presents a selection of quantitative results of our 90 bias tests. 

\subsection{Conceptualisation}

\paragraph{It's All Upstream\usecontext} 68\% (61 bias tests, Fig~\ref{fig:context}) address \textit{only} upstream LMs. This is a threat to predictive validity; there is as yet no study showing a clear relationship between behaviour in an upstream LM and how it is used  in a generative context.\footnote{Evidence of a weak connection was found in discriminative models \citep{goldfarb-tarrant-etal-2021-intrinsic, cao-2021-holistic}, we are unaware of comparable work for generative ones.} \citet{Chowdhery2022PaLMSL} acknowledge this concern: ``[W]hile we evaluate the pre-trained model here for fairness and toxicity along certain axes, it is possible that these biases can have varied downstream impacts depending on how the model is used.''\looseness=-1

Some bias tests clearly link bias in upstream LMs to harmful output in downstream tasks, such as in \citet{kurita-etal-2019-measuring}. However, references to downstream applications are often vague; authors rely on the unproven bias transfer hypothesis \cite{steed-etal-2022-upstream} to justify their approach, or mention downstream tasks in passing without clearly linking them to the way they have operationalised harm. 

\paragraph{What Biases Are We Measuring\harms and What Outcome Do We Want?\outcome } The literature struggles with specifying both biases---how it conceptualises bias, fairness, harm, etc.---and desired outcomes. 11\% of bias tests (Fig~\ref{fig:harm}) are not clear about the bias being studied, and 22\% (Fig~\ref{fig:outcome}) are not clear about the desired outcome (how a model would ideally behave), making \textit{unclear} the second most frequent choice for this attribute. Lack of clarity around bias conceptualisation is disappointing given this was the central message of the well-cited \citet{Blodgett_Barocas_2020}, and the papers we consider post-date its publication. 
The prevalence of unclear desired outcomes is also striking; we expected to find some fuzzy conceptualisations of bias, but were surprised that so much research is unclear on what behaviour a good model should have.\looseness=-1  

Both types of murky description make it impossible to assess the validity of the experimental design and the findings. Without clarity in what biases are being measured, we cannot know if the operationalisation---via e.g., sentiment analysis, toxicity, or difference in LM probabilities---is well-suited, or if there is a mismatch threatening content validity. \looseness=-1 For example, without defining the anticipated harm, it is unclear if comparing sentiment is an appropriate measure of that harm (as we found in \citet{schwartz-2021-ensemble, hassan-etal-2021-unpacking-interdependent}).
Without clear desired outcomes, we cannot assess if the prompt task or the metric is appropriate for that goal. If the desired outcome is to ensure that a model \textit{never} generates toxic content, 
both carefully handpicked prompts and automatically generated adversarial word salad are both likely to be helpful in accomplishing this goal, each with different limitations. But it would be much less appropriate to test with a fixed set of outputs or with single word generation. Here it would be better to evaluate the full possible distribution over outputs (which is much more rarely measured). If instead we desire that the model behaves acceptably in \textit{certain} contexts, then more constrained generation and evaluation may be both a reasonable and an easily controlled choice.

Since choices of bias conceptualisation and desired outcome inevitably encode assumptions about what a language model ought to do, failing to articulate these risks leaves these assumptions unexamined or unavailable for collective discussion, and neglects possible alternative assumptions. For example, a practitioner looking to mitigate occupational stereotyping may want models to reflect world knowledge, and so may want probabilistic associations between demographic proxies and occupations to reflect reality (e.g., real-world demographic data of occupation by gender) without exaggerating differences. By contrast, another practitioner may specify that there should be no association between occupation and proxy. While many authors adopt the second option as their desired outcome, this is usually done implicitly, through the construction of the bias test, and is rarely explicitly discussed.\looseness=-1 



\paragraph{Risks of Invariance \outcome} Many tests implicitly adopt invariance as a desired outcome, where a model should treat all demographic groups the same---e.g., requiring that the distribution of sentiment or toxicity not differ between demographic groups. This fails to take into account the effect of confirmation bias, whereby already stereotyped groups will be more affected by negative content due to people's propensity to recall confirmatory information \cite{nickerson1998confirmation}. This also neglects the group hierarchies that structure how different demographic groups experience the world; as \citet{hanna2020towards} put it, ``[G]roup fairness approaches try to achieve sameness across groups without regard for the difference between the groups....This treats everyone the same from an algorithmic perspective without acknowledging that people are not treated the same.''\looseness=-1

\paragraph{Stereotypes $\neq$ Negative Assumptions\harms}
Stereotypes form the majority of investigated harms (Fig~\ref{fig:harm}), but like \citet{blodgett-etal-2021-stereotyping}, we observed inconsistencies in how stereotypes are conceptualised. For example, some work conceptualises stereotypes as commonly held beliefs about particular demographic groups (and anti-stereotypes as their inverse) \citep{li-etal-2020-unqovering}, while others conceptualise stereotypes as negative beliefs \citep{zhou-etal-2022-sense, dinan-etal-2022-safetykit}, possibly conflating negative sentiment and stereotyping. 
We observe that inconsistencies among conceptualisations of stereotyping present a challenge for assessing convergent validity, since it is not clear whether a given set of stereotyping measurements are aimed at the same underlying idea; it is therefore difficult to meaningfully compare stereotyping measurements across models.



\subsection{Operationalisation}
\paragraph{Mind Your Origins \origin} For 66\% of bias tests (Fig~\ref{fig:origin}), prompts are either developed by the paper's authors, or else developed by authors of another paper and borrowed.\footnote{7 additional tests used author-created prompts with others.} Prompts are inevitably shaped by their authors' perspectives; while author-developed prompts can take advantage of authors' expertise, they also risk being limited by authors' familiarity with the biases under measurement.\footnote{This is made more problematic because these limitations are hidden, as authors rarely disclose their perspectives and backgrounds.} Few of these author-developed prompts were evaluated by other stakeholders; \citet{groenwold-etal-2020-investigating} is an encouraging exception, where prompt quality was assessed by annotators who are native speakers of African-American English or code-switchers. Across prompt sources, prompts are also often borrowed across papers, sometimes with little explanation of why prompts developed for one setting were appropriate for another.

\paragraph{Measuring Apples by Counting Oranges \measurement}
23 bias tests (26\%, Fig~\ref{fig:measure})
operationalise bias by checking whether generated text referencing marginalised groups yields lower sentiment than text not referencing such groups.
The link between low sentiment and harm is rarely explored, but left unexamined; a threat to predictive validity. 
Sentiment is often a poor proxy for harm; \citet{Sheng_Chang_Natarajan_Peng_2019} introduce the concept of \emph{regard} as a more sensitive measure of attitudes towards a marginalised group, observing that
sentences like \emph{GROUP likes partying} will yield positive sentiment but potentially negative regard. Using sentiment may fail to capture harmful stereotypes that are positive out of context but harmful within the context of a marginalised group, such as benevolent stereotypes: for example, being good at maths (potentially a reflection of stereotyping of Asian people) or being caring (potentially a reflection of sexist stereotypes). Many stereotypes 
have neutral valence (e.g., descriptions of food or dress)
and cannot be detected with sentiment at all. 

Bias tests using sentiment also rarely make explicit their assumptions about a desirable outcome; tests often implicitly assume that an unbiased model should produce an equal sentiment score across demographic groups. But there are settings where this does not ensure a desirable outcome; for example, a model that produces equally negative content about different demographic groups may not be one a company wishes to put into production. For some settings alternative assumptions may be appropriate---for example, requiring a model to produce positive content may be appropriate for a poetry generator \cite{Sheng_Uthus_2020} or for child-directed content---reinforcing the importance of evaluating language models in their contexts of use.\looseness=-1

\paragraph{My Model is Anti-Schoolgirl: Imprecise Proxies and Overreliance on Identity Terms \proxy \explicitdemo}
Bias tests exhibit surprisingly little variation in the demographic proxies they choose (Fig~\ref{fig:proxy}). Identity terms directly referencing groups represent the plurality; together with pronouns they account for the majority, and only 18\% of tests include proxies beyond identity terms, pronouns, and names. Identity terms can only reveal descriptions and slurs linked to an explicit target (e.g., \emph{a woman}, \emph{Muslims}). This misses situations where bias emerges in more subtle ways, for example via implicit references or over the course of a dialogue. 

We observe significant variation with regard to justifications for proxy terms; 71\% of tests fail to give reasoning for the demographic terms that they use, and 20\% fail even to \textit{list} the ones that they use, hampering our ability to evaluate content validity. Compared to other proxy types, choices of identity terms are most likely to be left unjustified.
For example, the description ``male indicating words (e.g., man, male etc.) or female indicating words (woman, female etc.)'' \cite{brown2020language} treats the concepts of ``male-indicating'' and ``female-indicating'' as self-evident, while \citet{dinan-etal-2020-multi} refer to ``masculine and feminine [] tokens.''\looseness=-1

Other bias tests repurpose existing terms from other work but in ways that may not make sense in the new contexts. For example, to represent religion (as a concept, not individual religious groups), one paper borrows the terms \textit{Jihad} and \textit{Holy Trinity} from \citet{nadeem-etal-2021-stereoset}. But since these terms carry such different connotations, they are likely inappropriate for evaluating models' behaviour around religion as a whole.
Another borrows \textit{schoolgirl} from \citet{bolukbasi}, who originally contrast the term with \emph{schoolboy} to find a gender subspace in a word embedding space. However, given its misogynistic or pornographic associations \cite{birhane2021multimodal}, uncritical usage of the term to operationalise gender threatens convergent validity (with other works on gender) and predictive validity (with downstream gender harms).
Elsewhere, \citet{bartl-leavy-2022-inferring} reuse the Equity Evaluation Corpus (EEC) from \citet{Kiritchenko_Mohammad_2018}, but exclude the terms \emph{this girl} and \emph{this boy} because ```girl' is often used to refer to grown women [but] this does not
apply to the word `boy'''; we encourage this kind of careful reuse.\looseness=-1 


\paragraph{Gender? I Hardly Know Her\gender \proxy \demo}
Gender is the most common demographic category studied in these tests (38\%, Fig~\ref{fig:demo}). Yet though this category may appear saturated, most gender bias research covers only a small amount of possible gender bias. An easy majority of work analyses only binary gender, and over half of this does not even \textit{acknowledge} the existence of gender beyond the binary, even with a footnote or parenthetical. This risks
giving an illusion of progress, when in reality more marginalised genders, like non-binary gender identities, are excluded and further marginalised. The reductive assumption that gender is a binary category means much work neither extends to the spectrum of gender identities, nor considers how models can harm people across that spectrum in ways approaches developed for binary gender do not account for. 

Across most gender bias work, discussions of the relationship between gender and proxy terms are missing or superficial; for example,  \emph{he} and \emph{she} are almost always described as male and female pronouns, though they are widely used by nonbinary individuals\footnote{\url{https://www.gendercensus.com/results/2022-worldwide/\#pronouns}} \cite{Dev_Monajatipoor_Ovalle_Subramonian_Phillips_Chang_2021} 
(an exception is \citet{Munro_Morrison_2020}, who write of ``people who use `hers,' `theirs' and `themself' to align their current social gender(s) with their pronouns' grammatical gender''). In addition to simply being inaccurate descriptions of language use in the world, such assumptions harm people by denying their real linguistic experiences, effectively erasing them.
Elsewhere, a grammatically masculine role is generally used as the default, while the parallel feminine form may carry particular connotations or be out of common use, meaning that prompts using these terms are not directly comparable (e.g., \emph{poet} vs. \emph{poetess}).\looseness=-1 

\paragraph{Well Adjusted? \measurement}
35 tests (Fig~\ref{fig:measure}) operationalise bias by comparing the relative probability of proxies in sentences about different topics. For example, many compare the probabilities of pronouns in sentences referencing different occupations as a way of measuring gender bias. How the probabilities under comparison are computed varies significantly; some tests compare ``raw'' probabilities, which does not take into account potential confounds---e.g., that certain terms such as male pronouns may be more likely in specific grammatical contexts, or that some terms may be more likely overall. Others use adjusted or normalised probabilities \cite{ahn-oh-2021-mitigating, kurita-etal-2019-measuring}, which carry their own risk of being less similar to real-world language use, potentially threatening the test's ecological validity. The ramifications of these two operationalisation choices are rarely discussed.\looseness=-1 


\subsection{Basic Details \& Scope}
\paragraph{Narrow Field of View\model}
We find that most bias tests investigate few models.
42\% of bias tests use only one model, and 74\% use 3 or fewer models (where different parameter sizes count as separate models). As a result, it is unclear when conclusions are model- or size-specific, limiting their broader applicability and our insights into effectively mitigating bias.\looseness=-1

\paragraph{Speak English, Please.\langs} 87\% of bias tests examine only English (78), and of the 12 remaining that consider other languages, only two test in a language that is not highly resourced. Among tests beyond English, we identify two predominant types. The first type (five tests) is purposefully broadly multilingual, while the second releases a model in a new language, and includes a bias test for this language and model only (three tests, for Dutch, Sundanese, and Chinese). PaLM \cite{Chowdhery2022PaLMSL}, a massively multilingual model, tests bias only in English, even though English bias measurements are unlikely to apply universally. 

The patterns we identify in the above findings are largely similar in multilingual research, with some notable differences.\footnote{ Appendix~\ref{app:multi_stats_all} contains graphs for multilingual studies.} The reliance on only upstream LMs is exacerbated, with only one paper considering use in a downstream task \cite{mi2022pangubot}. No bias tests express \textit{no impact of demographic term} as a desired outcome, suggesting that counterfactuals are less popular in multilingual research. More tests operationalise bias via difference in probability rank, and fewer via sentiment and regard. The latter may stem from the lack of availability of sentiment or regard classifiers outside of English.\looseness=-1

\paragraph{A Bender Rule for Cultural Contexts} 
Most English bias tests assume an American or Western  context (a general trend in NLP \cite{bhatt-etal-2022-contextualizing}). Although the appropriateness of demographic group and proxy choices unavoidably depend on cultural context, assumptions about such context are rarely explicitly stated; exceptions include \citet{Li_Khashabi_Khot_Sabharwal_Srikumar_2020} and \citet{Smith_Williams_2021}.\looseness=-1 

\section{Discussion}
\paragraph{Validity and Reliability} Whereas validity asks, ``Is [the measurement] right?'', \emph{construct reliability} asks, ``Can it be repeated?'' \cite{quinn2010analyze}. 
Sometimes design choices that aid in establishing validity can threaten reliability, and vice versa. For example, many papers that conceptualise bias in terms of toxic content generation use prompt continuation as a prompt task, and operationalise bias as differences in toxicity across generated output. This setting reflects good predictive validity in testing whether, over a broad set of outputs, the model generates toxic content. However, reliability may be threatened, as the test is brittle to choices such as decoding parameters \citep{akyurek-etal-2022-challenges}. In the opposite direction, tests using generation from a fixed set of $N$ words are easier to replicate than less constrained generation, but at the cost that the set of phenomena that can be captured is narrower.

Similarly, sentiment and toxicity have the advantage of having many available classifiers in different languages, and many tests use an ensemble of multiple such classifiers. Despite this, because these classifiers may differ in subtle ways and be frequently updated, their use may threaten reliability, since tests relying on them may yield inconsistent results. By contrast, \emph{regard} is operationalised via a classifier developed by \citet{Sheng_Chang_Natarajan_Peng_2019}, and as papers' domains diverge from what Sheng et al. intend, validity is increasingly threatened. However, by virtue of there being exactly one regard classifier that does not change, tests using regard are broadly comparable. Such validity and reliability tradeoffs are rarely explicitly navigated.

\paragraph{Unknown Unknowns} 
Our taxonomy is a reflection of what is missing as much as what is present. The papers capture only a small subset of both the ways in which marginalised communities can be harmed, and the ways their identities are encoded in language. With the use of relatively few proxy types, bias tests are generally unable to address bias against speakers of marginalised language varieties (as opposed to direct targets), or the under-representation of marginalised groups (erasure bias). 



\section{Recommendations} 
\label{sec:guidance}
Guided by our analysis, we formulate the following list of questions that future bias research can consult to inform experimental design. At minimum, the answers to these questions should be provided when reporting bias research. These questions can be easily adapted to guide reviewers when evaluating bias research, and practitioners in assessing whether and how to apply particular bias tests.\looseness=-1

\textbf{Scope}
\begin{itemize}
    \item \textbf{More than the bare minimum \langs \model}
    If releasing a multilingual model, have you tested for bias across multiple languages, beyond English?
    \item \textbf{All of Sesame Street \model}
    Why are you testing these particular models? Can your test be adapted to other models? 
\end{itemize}

\textbf{Conceptualisation
}\begin{itemize}
    \item \textbf{Tell me what you want (what you really really want)} \outcome
    What is your desired model outcome, and how does your test allow you to measure deviation from that desired outcome? How does this outcome connect to your harm?
\end{itemize}

\textbf{Operationalisation
}\begin{itemize}
    \item \textbf{Make the implicit explicit\proxy \explicitdemo} Why are your chosen terms suitable proxies for the demographic groups you are studying? 
    What is the cultural context to which these terms are relevant?
    \item \textbf{Well-spoken \proxy}
    Have you considered the many ways a group identity can manifest linguistically?\looseness=-1
    \item \textbf{Don't reinvent the wheel \proxy \measurement} Did you consider relevant work from linguists and social scientists when designing your bias measures?
    \item \textbf{Broaden your horizons \demo}
    Can your work be expanded to further cultural contexts? 
    Is a binary conceptualisation of gender appropriate, or necessary?

\end{itemize}

\textbf{Other Validity Considerations
}\begin{itemize}
    \item \textbf{Consider the future}
    Does your test allow us to make predictions about downstream behaviour (predictive validity)?
    \item \textbf{Do a reality check}
    Does your measurement approach reflect ``real world'' language and model usage (ecological validity)?
    \item \textbf{Beware of collateral damage} Can your measurement approach cause harm or other impacts (consequential validity)?
\end{itemize}



\section{Conclusion}

We hope that via our taxonomy and analysis, practitioners are better-equipped to understand and take advantage of the wealth of emerging approaches to bias testing---in particular, to clearly conceptualise bias and desired model outcomes, design meaningful and useful measurements, and assess the validity and reliability of those measurements. 

\section{Limitations}
Our search was conducted exclusively in English, and we may have missed relevant papers written in other languages; this may have influenced the heavy English skew in our data.

Some of the annotations of attributes and choices in this taxonomy rely on subjective judgements, particularly with regards to the clarity of conceptualisations of bias, desired outcomes, and justifications of proxy choices. As with any qualitative work, these results are influenced by our own perspectives and judgement. We did our best to address this through regular discussion, identifying disagreements early on when designing the taxonomy, and adopting a ``generous'' approach. 

\section{Ethics Statement}


All measurement approaches discussed in this paper encode implicit assumptions about language and culture, or normative assumptions about what we ought to do, which must be made explicit for them to be properly evaluated. We acknowledge our work will have been shaped by our own cultural experiences, and may similarly encode such assumptions.


\section*{Acknowledgements}
We would like to thank our anonymous reviewers
for their feedback. Eddie L. Ungless is supported
by the UKRI Centre for Doctoral Training in Natural Language Processing, funded by the UKRI
(grant EP/S022481/1) and the University of Edinburgh, School of Informatics.

\bibliography{anthology,custom}
\bibliographystyle{acl_natbib}

\clearpage
\appendix

\section{Appendix}
\label{sec:appendix}

\subsection{Types of Validity}\label{ss:validity}
See Table \ref{tab:threats_examples}.
\input{tables/threats_table}

\subsection{Full Taxonomy} \label{ss:full_taxonomy}

We provide here details of our taxonomy (Table \ref{tab:taxonomy}), including detailed explanations of each option.

\paragraph{Language(s)} What language(s) is/are investigated?

\paragraph{Model(s)} What model(s) is/are investigated?

\paragraph{Code available?} Is code for the proposed bias test publicly available?
\begin{itemize}
\setlength{\itemsep}{0pt}
\item yes/no
\end{itemize}

\paragraph{Use context} What context will the model be used in?
\begin{itemize}
\setlength{\itemsep}{0pt}
\item zero-shot/few-shot
\item upstream LM
\item dialogue
\item Q\&A
\end{itemize}

\paragraph{Bias conceptualisation} How is bias---bias, fairness, stereotypes, harm, etc.---conceptualised?
\begin{itemize}
\setlength{\itemsep}{0pt}
\item stereotyping: paper identifies stereotyping as a harm
\item toxic content generation: paper identifies negative or toxic (including racist, sexist, etc. ) content as a harm
\item other: paper identifies something else as a harm (annotator includes description in a comment)
\item unclear: it is unclear how the paper conceptualises bias or harm
\end{itemize}

\paragraph{Prompt task} What is the prompt task?
\begin{itemize}
\setlength{\itemsep}{0pt}
\item sequence scoring: model is tasked with scoring various sequences
\item single word generation: model is tasked with generating a single word
\begin{itemize}
    \item analogy: model is tasked with completing an analogy
\end{itemize}
\item prompt continuation: model is tasked with continuing a prompt (2+ words)
\item full sentence response: model is tasked with responding to a full sentence
\end{itemize}

\paragraph{Prompt origin} Where do the prompts originate?
\begin{itemize}
\setlength{\itemsep}{0pt}
\item author: prompts are written by the author, or sourced from a paper where they are written by that paper's authors
\item crowd-sourced: prompts are crowd-sourced from workers other than the paper authors, or sourced from a paper where they are crowd-sourced
\item corpus: prompts are scraped from a corpus, including Wikipedia or social media, or sourced from a paper where they are scraped from a corpus
\item automatically generated: prompts are generated by a model
\end{itemize}

\paragraph{Metric} What metric or strategy is used to measure bias or harm?
\begin{itemize}
\setlength{\itemsep}{0pt}
\item output content assessed: assessment of output content, e.g., presence of stereotypes
\item output quality assessed: mentions of demographic groups lead to differences in quality of output content, e.g., grammaticality or relevance
\item difference in probability (ranking over fixed set): which of a fixed set of options is more probable
\item most probable option(s): assess the top 1 or N options
\item difference in output distributions: assessment of entire output distributions under different conditions
\item difference in regard: mentions of demographic groups lead to differences in regard of output content
\item difference in sentiment: mentions of demographic groups lead to differences in sentiment of output content
\item difference in toxicity: mentions of demographic groups lead to differences in toxicity of output content
\end{itemize}


\paragraph{Desired outcome} How is a good model outcome conceptualised?
\begin{itemize}
\setlength{\itemsep}{0pt}
\item no impact of demographic term(s): mentions of demographic groups do not change model predictions. 
\item negative stereotype not in model: mentions of demographic groups do not result in output reflecting stereotypes
\item other: another conceptualisation (annotator includes description in comment)
\item unclear: it is unclear how the paper conceptualises a good model outcome
\end{itemize}

\paragraph{Demographics} For which demographic groups is bias or harm investigated?
\begin{itemize}
\setlength{\itemsep}{0pt}
\item gender
\item ethnicity/race
\item religion
\item sexual orientation
\item other: other demographic groups (annotator includes description in comment)
\end{itemize}

\paragraph{Proxy type(s)} Which term(s) is/are used to proxy the demographic groups under investigation?
\begin{itemize}
\setlength{\itemsep}{0pt}
\item identity terms: terms that refer directly to demographic groups, such as \emph{Muslim}
\item pronouns
\item names: people's names
\item roles: terms that refer to social roles, such as \emph{mother}
\item dialect features: terms reflecting dialectal variation, such as lexical items associated with African American Language (AAL)
\item other: other terms (annotator includes description in comment)
\item unclear: it is unclear what terms are used
\end{itemize}

\paragraph{Explicit demographics} Are the choices of demographic groups and accompanying proxies clearly defined and explained?
\begin{itemize}
\setlength{\itemsep}{0pt}
\item yes/no
\end{itemize}

\paragraph{Gender scope} For work investigating gender, how is gender treated?
\begin{itemize}
\setlength{\itemsep}{0pt}
\item binary gender only: gender is treated as binary, specifically man and woman, or male and female 
\item binary gender only plus acknowledgement: gender is treated as binary, accompanied by an acknowledgement that gender is not binary
\item binary and other genders: gender treatment includes men, women and other marginalised genders
\item other genders only: gender treatment excludes binary genders 
\end{itemize}

\subsection{Results from Taxonomy for Multilingual and Non-English Bias Tests}
\label{app:multi_stats_all}
\input{figures/multi_stats}

\clearpage

\end{document}

%% file: tables/taxonomy_table_bigger.tex
\begin{table*}[t]
    \footnotesize
    \def\arraystretch{1.1}
    \setlength{\tabcolsep}{0.4em}
    \centering
    \begin{tabular}{@{}p{2.8cm}p{5cm}p{7.8cm}@{}} \toprule
    {\bf Attribute} & {\bf Description} & {\bf Choices} \\ \hline
    \multicolumn{3}{@{}l}{\bf Basic details and scope} \\ 
    Language(s)\langs & What language(s) is/are investigated? & open-ended \\
    Model(s)\model & What model(s) is/are investigated? & open-ended \\ 
    Code available? & Is code for the proposed bias test publicly available? & yes, no \\ \cmidrule{1-3}
    \multicolumn{3}{@{}l}{\bf Conceptualisation} \\ 
    Use context\usecontext & What context will the language model be used in? & zero-shot/few-shot, upstream LM, dialogue, Q\&A \\
    Bias conceptualisation \harms & How is bias---bias, fairness, stereotypes, harm, etc.---conceptualised? & stereotyping, toxic content generation, other, unclear \\ 
    Desired outcome\outcome & How is a good model outcome conceptualised? & no impact of demographic term(s), negative stereotype is not in model, no harmful output generated, other, unclear \\ \cmidrule{1-3}
    \multicolumn{3}{@{}l}{\bf Operationalisation} \\ 
    Prompt task\task & What is the prompt task? & sequence scoring, single word generation, prompt continuation, full sentence response \\
    Prompt origin\origin & Where do the prompts originate? & author, crowd-sourced, corpus, automatically generated \\
    Metric \measurement & What metric or strategy is used to measure bias or harm? & output content assessed, output quality assessed, difference in probability (ranking over fixed set), most probable option(s), difference in output distributions, difference in regard, difference in sentiment, difference in toxicity \\
    Demographics\demo & For which demographic groups is bias or harm investigated? & gender, ethnicity/race, religion, sexual orientation, other \\
    Proxy type(s)\proxy  & What term(s) is/are used to proxy the demographic groups under investigation? & identity terms, pronouns, names, roles, dialect features, other, unclear \\
    Explicit demographics\explicitdemo & Are the choices of demographic groups and accompanying proxies clearly defined and explained? & yes, no \\
    Gender scope\gender & For work investigating gender, how is gender treated? & binary gender only, binary gender only plus acknowledgement, binary and other genders, other genders only \\

    \bottomrule
    \end{tabular}
    \caption{Our taxonomy of attributes. We provide full descriptions of each attribute's options in the appendix (\ref{ss:full_taxonomy}).}
    \label{tab:taxonomy}
\end{table*}

%% file: figures/overall_stats.tex

\begin{figure*}[ht]
    \centering   
    \begin{subfigure}[b]{0.31\textwidth}
        \includegraphics[width=\textwidth]{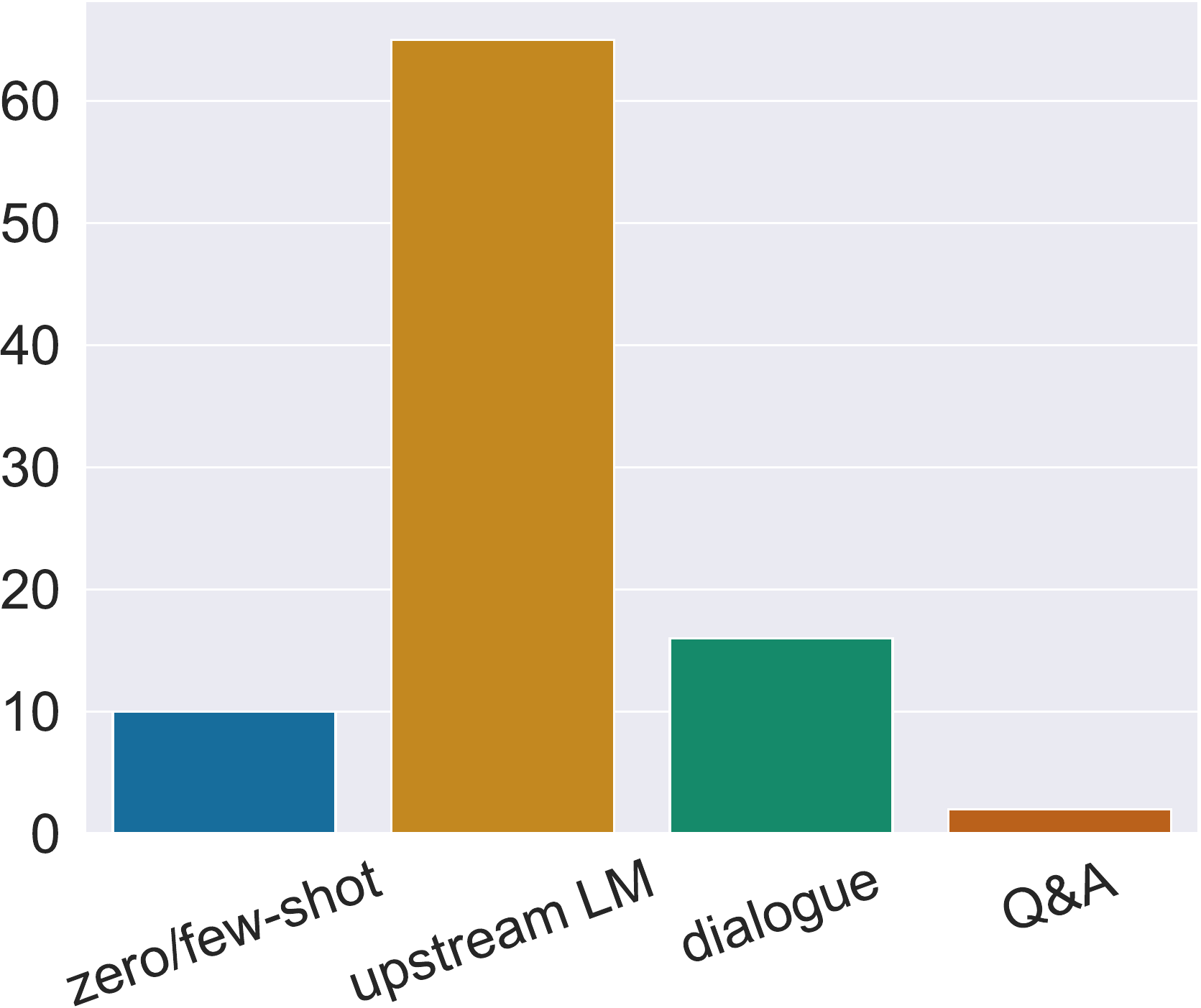}
        \caption{Use Context}
        \label{fig:context}
    \end{subfigure}
    \begin{subfigure}[b]{0.31\textwidth}
        \includegraphics[width=\textwidth]{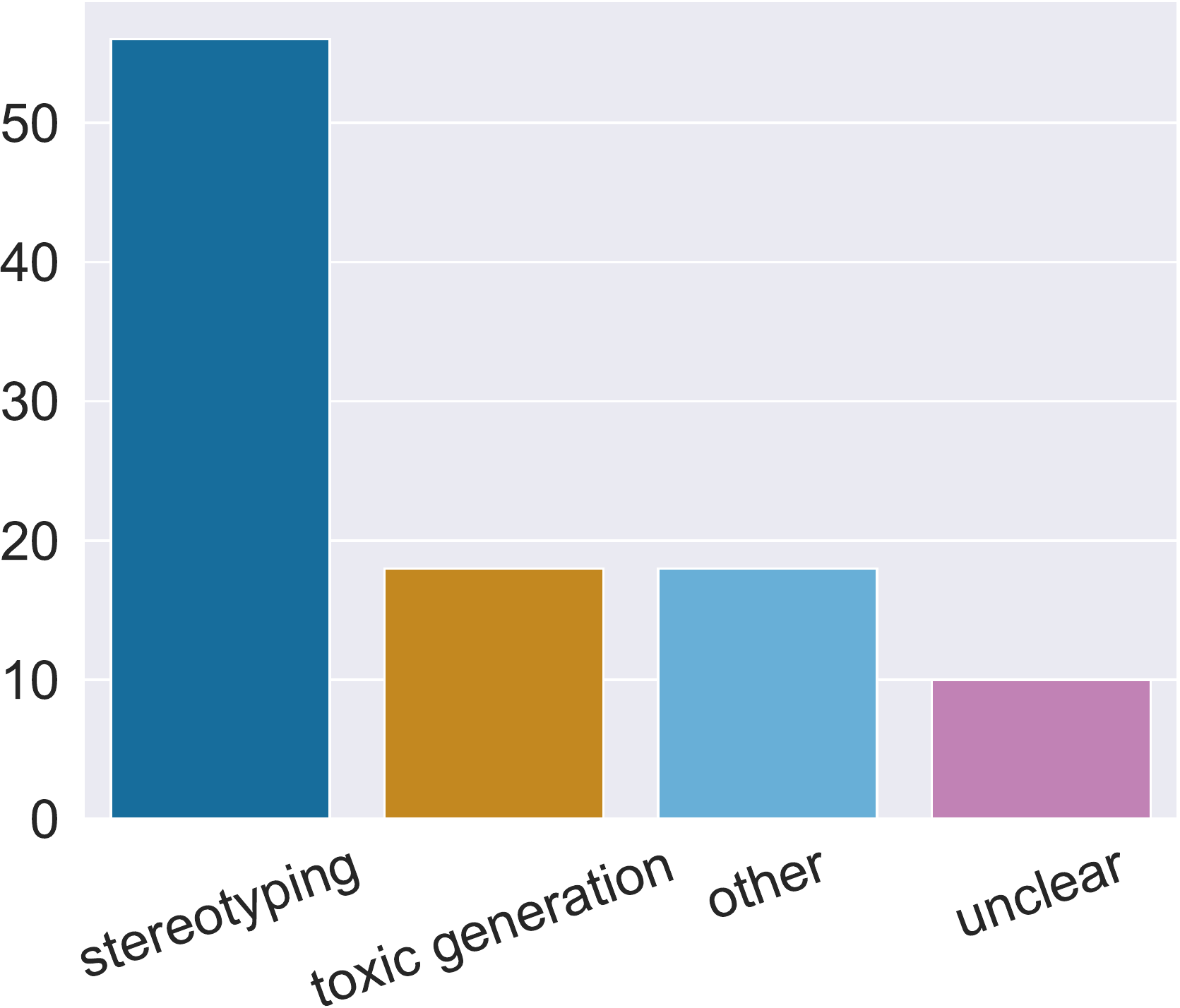}
        \caption{Biases/Harms}
        \label{fig:harm}
    \end{subfigure}
    \begin{subfigure}[b]{0.31\textwidth}
        \includegraphics[width=\textwidth]{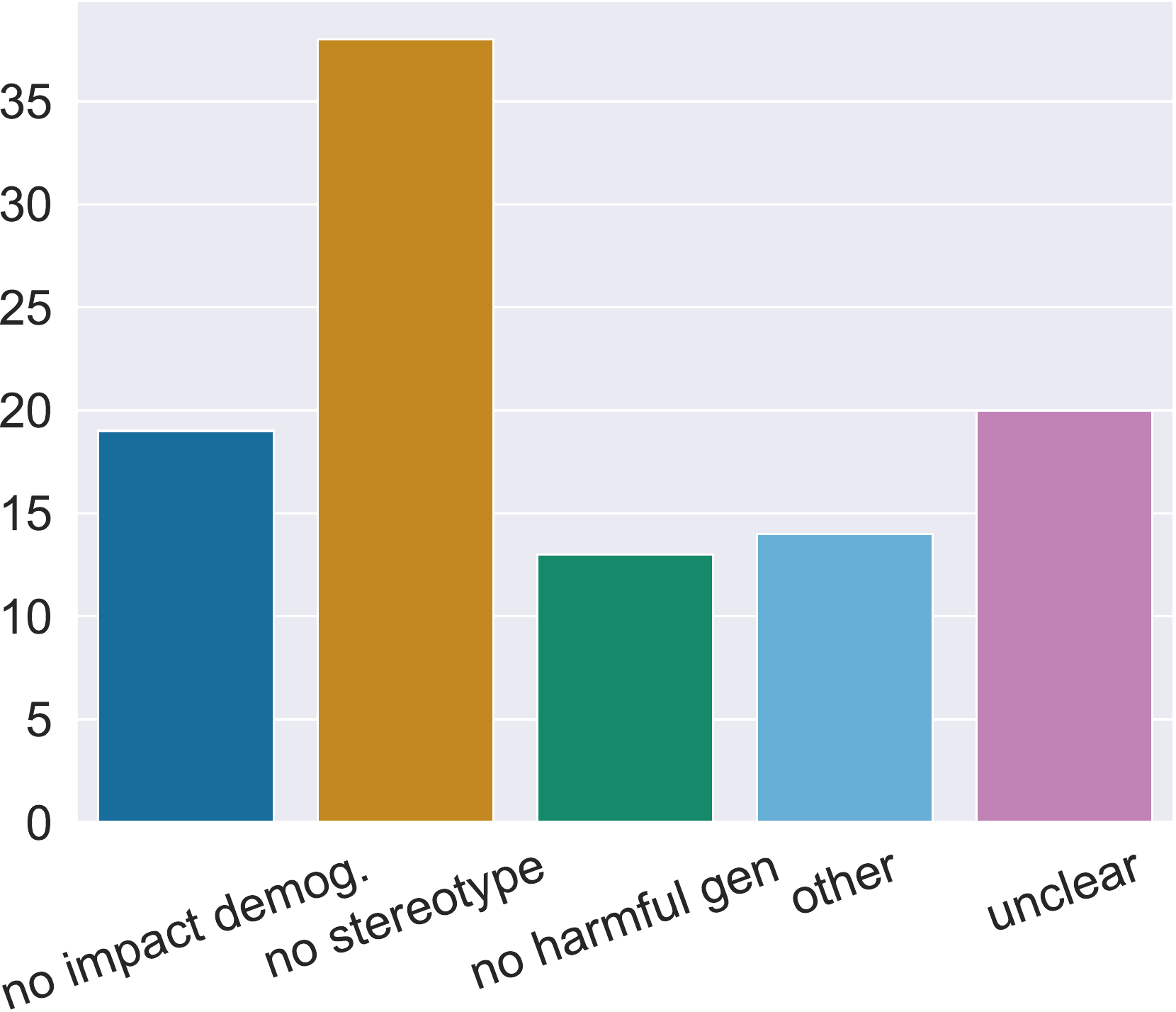}
        \caption{Desired Outcome}
        \label{fig:outcome}
    \end{subfigure}
    ~ 
        \begin{subfigure}[b]{0.31\textwidth}
        \includegraphics[width=\textwidth]{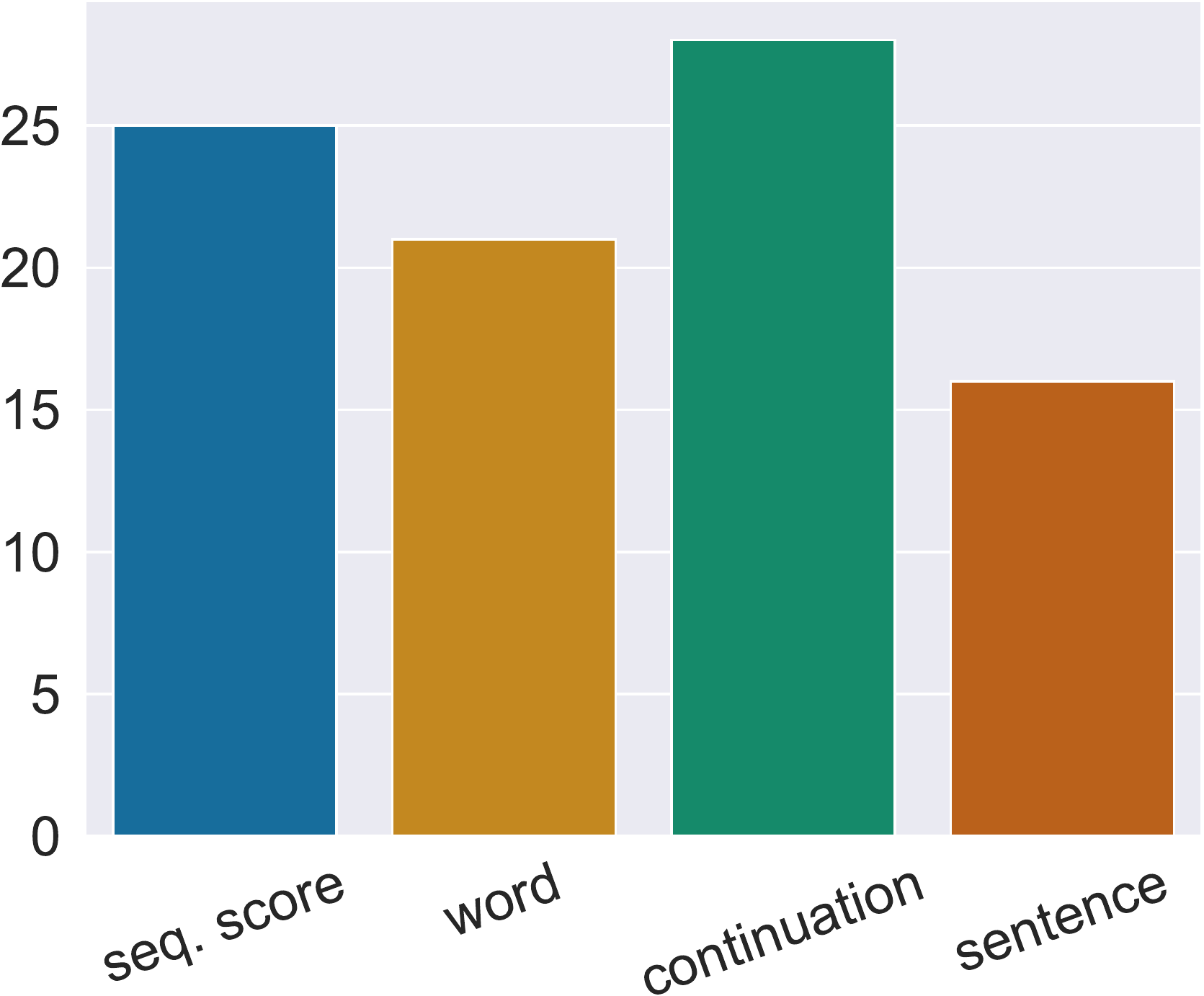}
        \caption{Prompt Task}
        \label{fig:task}
    \end{subfigure}
    \begin{subfigure}[b]{0.31\textwidth}
        \includegraphics[width=\textwidth]{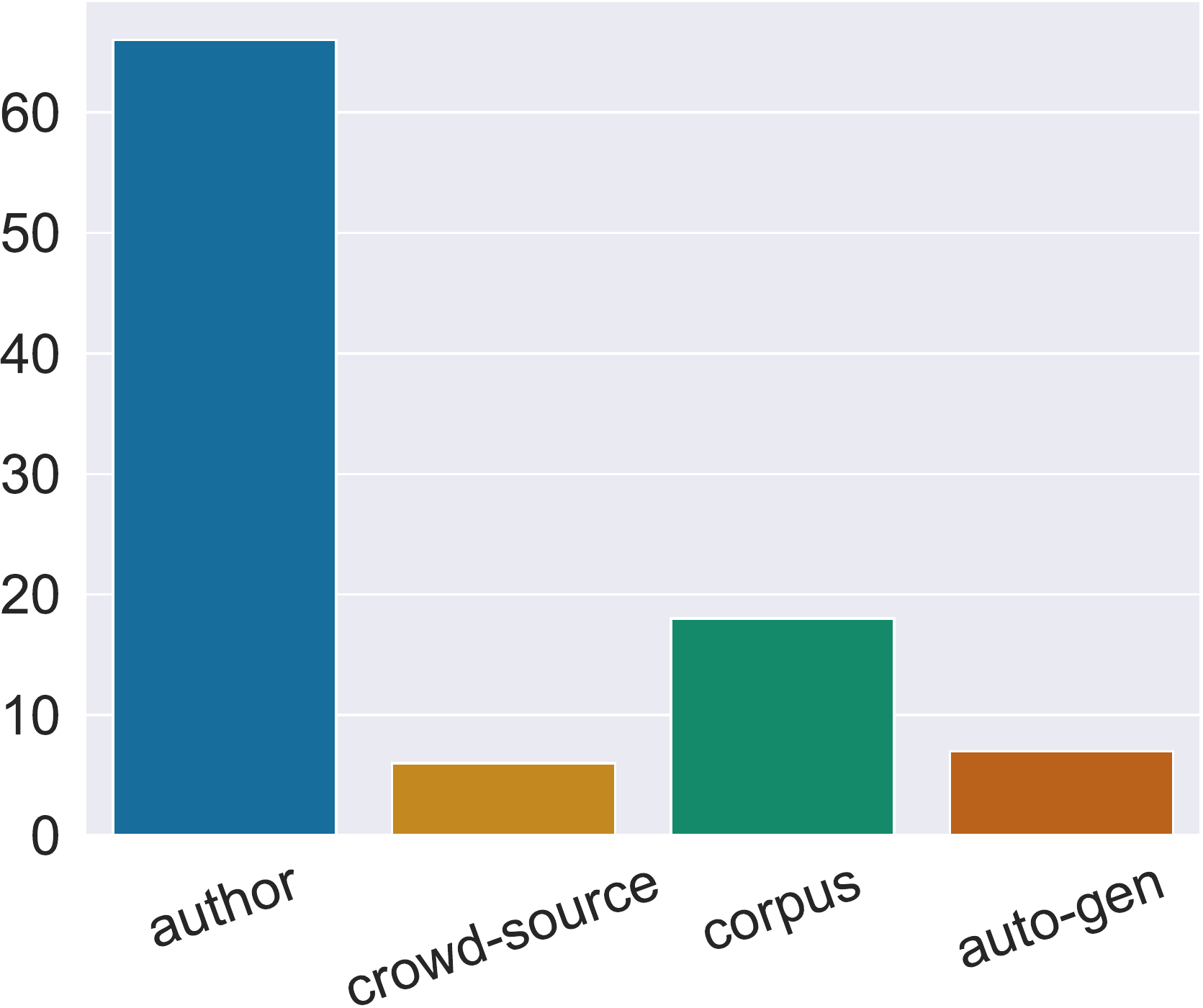}
        \caption{Prompt Origin}
        \label{fig:origin}
    \end{subfigure}
    \begin{subfigure}[b]{0.335\textwidth}
        \includegraphics[width=\textwidth]{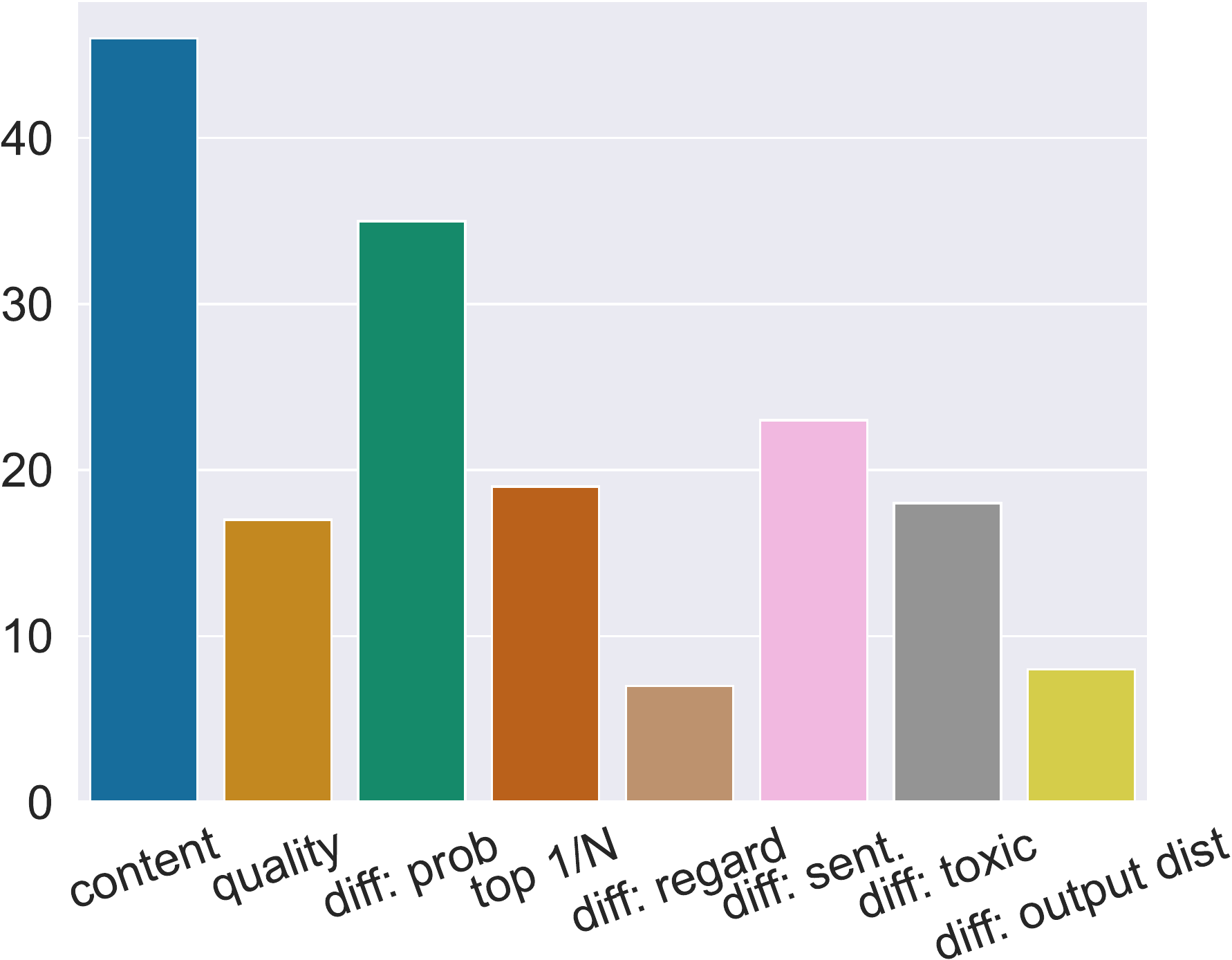}
        \caption{Metric}
        \label{fig:measure}
    \end{subfigure}
    ~ 
        \begin{subfigure}[b]{0.31\textwidth}
        \includegraphics[width=\textwidth]{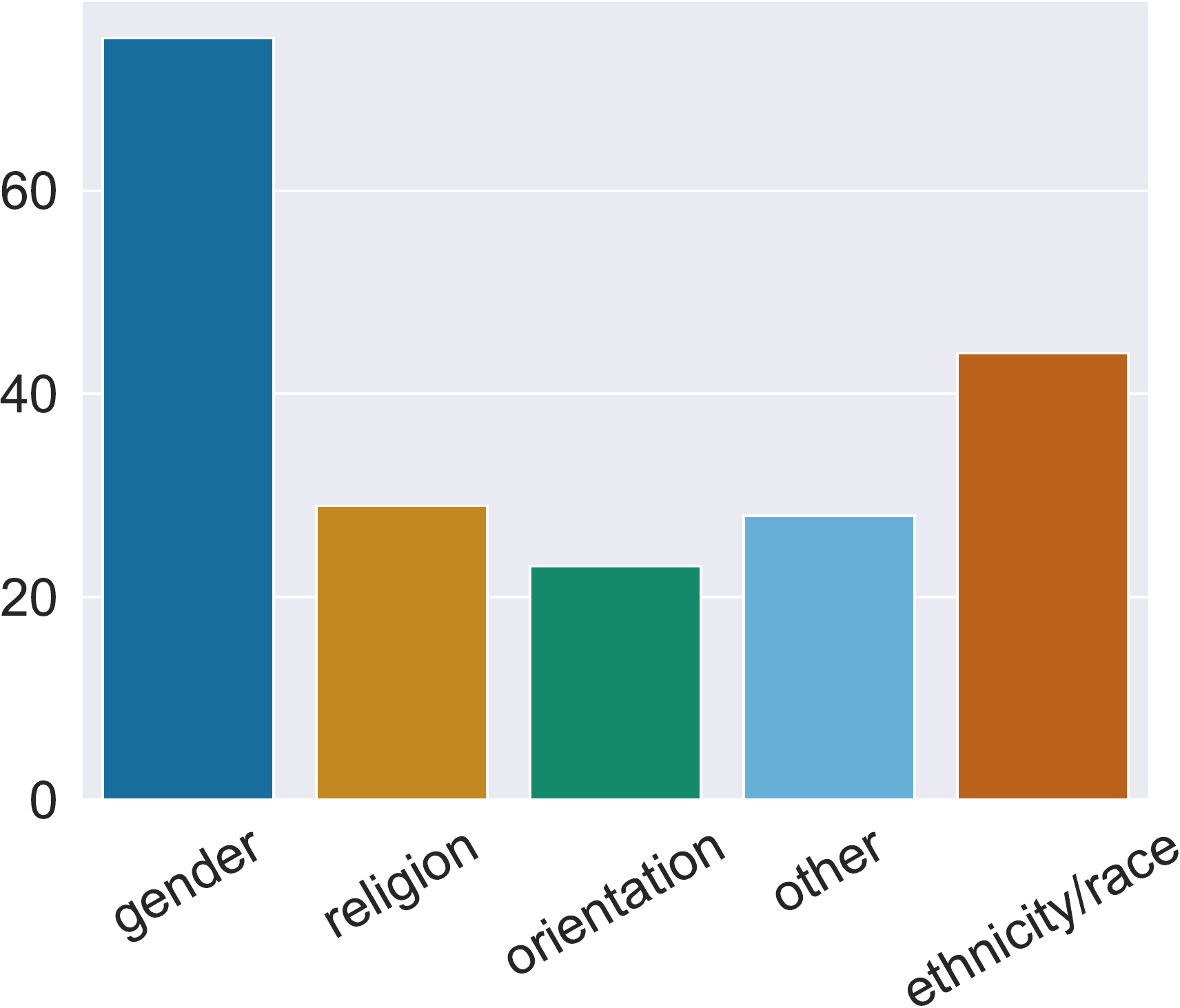}
        \caption{Demographic}
        \label{fig:demo}
    \end{subfigure}
    \begin{subfigure}[b]{0.335\textwidth}
        \includegraphics[width=\textwidth]{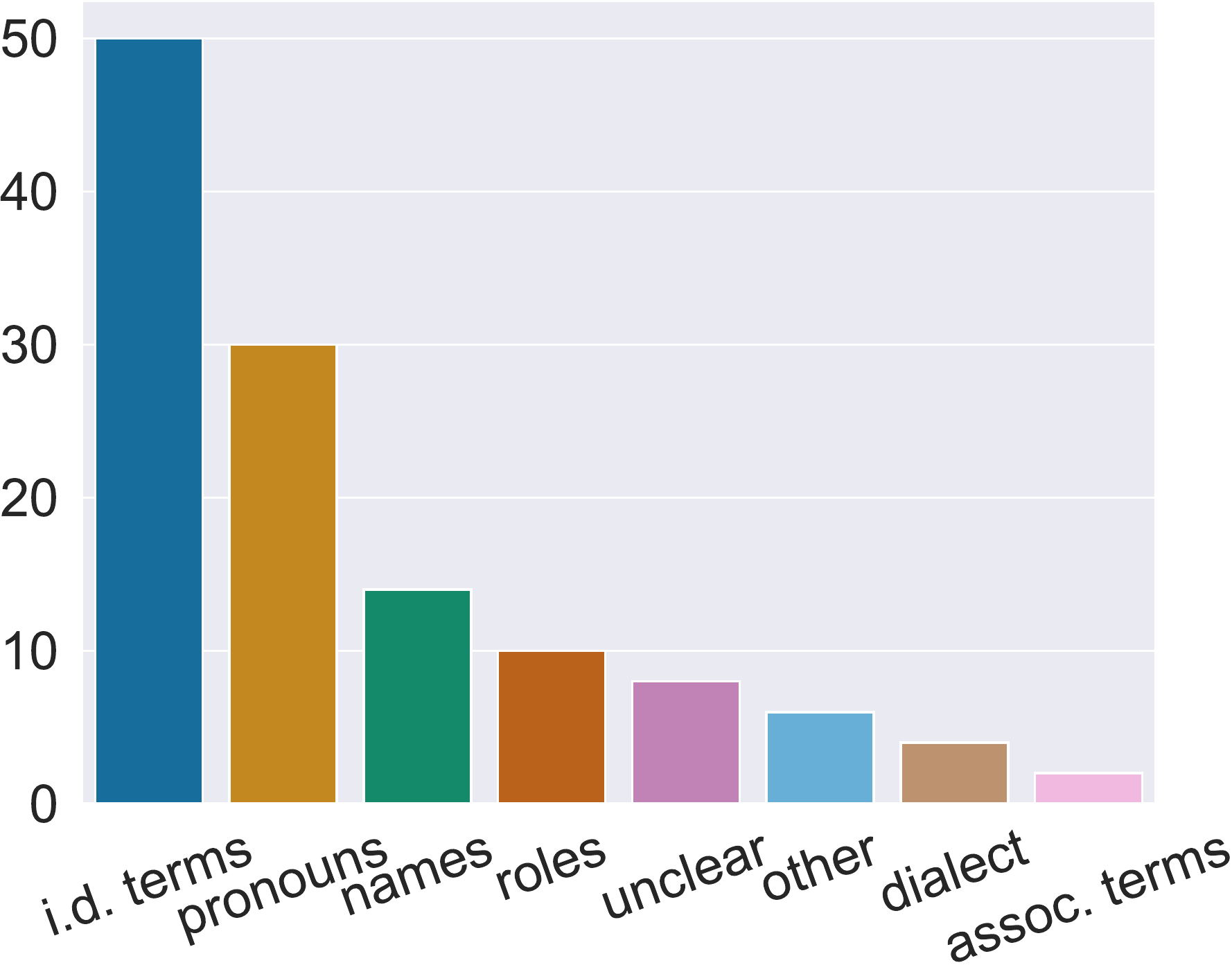}
        \caption{Demographic Proxy}
        \label{fig:proxy}
    \end{subfigure}
    \begin{subfigure}[b]{0.32\textwidth}
        \includegraphics[width=\textwidth]{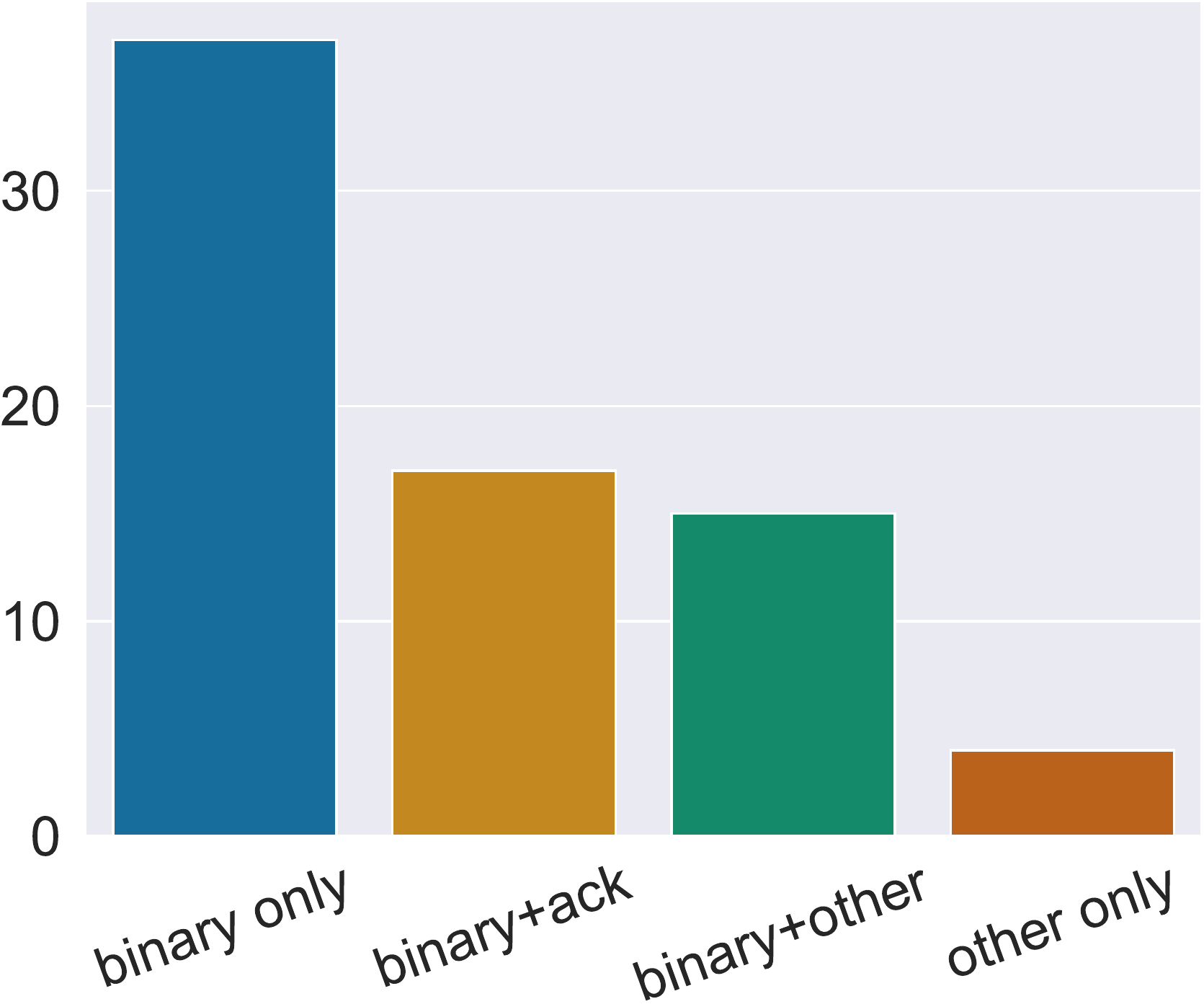}
        \caption{Gender Scope}
        \label{fig:gender}
    \end{subfigure}

    \caption{Our taxonomy (Table~\ref{tab:taxonomy}) applied to 90 bias tests. Full details of terminology in Appendix \ref{ss:full_taxonomy}.} 
    \label{fig:overall_stats}
    \vspace{-1em}
\end{figure*}

%% file: tables/threats_table.tex
\begin{table*}[ht]
    \centering
    \begin{tabular}{p{3.5cm}p{3cm}p{8.5cm}} 
    \toprule
        \textbf{Type of Validity} & \textbf{Short Definition} &\textbf{Example Threat} \\
        \midrule
        \textbf{Construct validity} && \\
        \hfill Face validity & Plausibility & Using BLEU score to measure relevance of generation - BLEU does not measure meaning \\
        \hfill Content validity & Effective operationalisation & Paper aims to measure fairness but results not split by demographic, unclear if some groups disproportionately affected \\
        \hfill Convergent validity & Correlation with existing measures & Proposed measures rarely compared to existing measures\\
        \hfill Predictive validity & Can predict related measurements & Authors assume upstream bias predicts downstream bias; this has not been proven 
        \\
        \hfill Consequential validity & Impact on world \& behaviours & People may assume low bias in LM will ensure low bias in finetuned model and feel ``safe'' using these models\\
        \hline
        \textbf{Ecological validity} & Results generalise to the world & By factoring out confounds on relative probabilities, measurement does not reflect typical use of model \\
    \bottomrule
    \end{tabular}
    \caption{Overview of threats to validity. Each threat is derived from examples found in our analysis.  }
    \label{tab:threats_examples}
    \vspace{-1em}
\end{table*}

%% file: figures/multi_stats.tex
\begin{figure*}[ht]
    \centering
    \includegraphics[width=\textwidth]{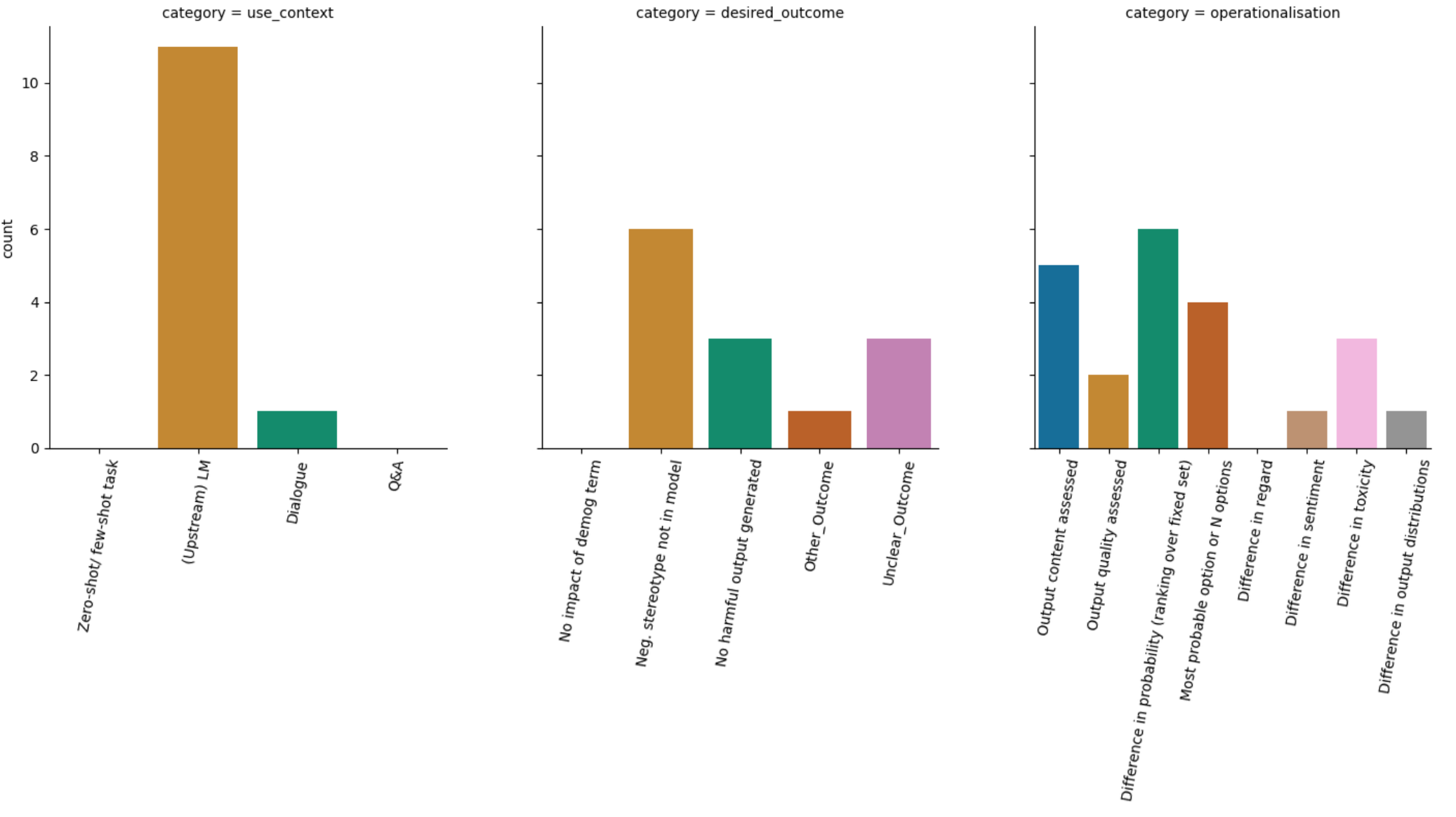}
    \caption{The same as Table \ref{tab:taxonomy}, isolated to the 12 multilingual bias tests to show the patterns there that differ from overall ones.}
    \label{fig:multi_stats}
    \vspace{-1em}
\end{figure*}